\documentclass[11pt]{article}
\usepackage[utf8]{inputenc}
\pdfoutput=1
\usepackage{amsmath,amsbsy,amsfonts,amssymb,amsthm,dsfont,fullpage,units}
\usepackage{booktabs,todonotes} 

\usepackage[colorlinks]{hyperref}
\usepackage{graphicx, multirow, xcolor}
\usepackage{comment}
\usepackage{bbm}
\usepackage{multirow}
\usepackage[ruled,noline, noend]{algorithm2e}
\graphicspath{{figures/}}

\newcommand{\says}[3]{\todo[size=\small,color=#2,inline]{#1 says: #3}}
\newcommand{\moustafa}[1]{\says{Moustafa}{red}{#1}}
\title{CAAD 2018: Generating Transferable Adversarial Examples}
\usepackage{soul}

\author{Yash Sharma \\ \texttt{ysharma1126@gmail.com} \and Tien-Dung Le \\ \texttt{tien.dung.le.ext@proximus.com}  \and
Moustafa Alzantot \\ \texttt{malzantot@ucla.edu}}

\begin{document}

\maketitle

\begin{abstract}
Deep neural networks (DNNs) are vulnerable to adversarial examples, perturbations carefully crafted to fool the targeted DNN, in both the \textit{non-targeted} and \textit{targeted} case. In the non-targeted case, the attacker simply aims to induce misclassification. In the targeted case, the attacker aims to induce classification to a specified target class. In addition, it has been observed that strong adversarial examples can transfer to unknown models, yielding a serious security concern.  

The NIPS 2017 competition was organized to accelerate research in adversarial attacks and defenses, taking place in the realistic setting where submitted adversarial attacks attempt to transfer to submitted defenses. The CAAD 2018 competition took place with nearly identical rules to the NIPS 2017 one. Given the requirement that the NIPS 2017 submissions were to be open-sourced, participants in the CAAD 2018 competition were able to directly build upon previous solutions, and thus improve the state-of-the-art in this setting. 

Our team participated in the CAAD 2018 competition, and won \textbf{1st} place in both attack subtracks, \textit{non-targeted} and \textit{targeted} adversarial attacks, and \textbf{3rd} place in \textit{defense}. We outline our solutions and development results in this article. We hope our results can inform researchers in both generating and defending against adversarial examples.
\end{abstract}

\section{Introduction}

An adversarial example is a perturbed input example, where the perturbation is carefully crafted to induce a desired behavior while remaining virtually imperceptible~\cite{szegedy2013intriguing,goodfellow2014explaining}. Such examples have been motivated in the image modality but are not limited to it, for example, attack algorithms have also been proposed for text and speech~\cite{carlini2018audio,alzantot2018generating}. In the \textit{non-targeted} case, the desired behavior would be to induce misclassification, classification to any label other than the true label. In the \textit{targeted} case, the desired behavior would be to induce classification to a chosen target class $t$. Though executing a successful targeted attack is obviously more difficult, especially on tasks where there are many similar classes (e.g. ImageNet~\cite{deng2009imagenet}), it has been observed that not only can both non-targeted and targeted adversarial examples be crafted against a targeted DNN, but these examples can transfer, induce a similar effect, to unknown models not seen during generation~\cite{liu2016delving, carlini2017towards,ead,ead_madry,ead_feature}. This is a serious security concern, as it suggests that even if your model cannot be accessed by the outside world, it can still be fooled by an unknown adversary. 

Defenses have been proposed for handling adversarial examples, though success has been limited. It has been shown that numerous proposed defenses achieve their apparently successful results through manipulating what existing attacks rely upon, and if that is circumvented, no robustness benefit is yielded. For example, in~\cite{athalye2018obfuscated}, it was shown that many recently proposed defenses, published in ICLR 2018, relied upon manipulating the gradients used by white-box attackers. If an attacker were to not use the gradient, they would circumvent the defense, and if an attacker was aware that this \textit{gradient obfuscation} was taking place, they could readily undo the manipulation and easily attack the targeted DNN~\cite{athalye2018obfuscated,chen2017zoo,alzantot2018genetic}. The one defense which has stood the test of time is adversarial training~\cite{szegedy2013intriguing, madry, tramer2017ensemble}, training upon a large number of adversarial examples to attain robustness, though this solution is limited by 1) not offering any guarantees, 2) requiring significantly more expensive training, which can be impractical on high-dimensional datasets (e.g. ImageNet), and 3) tending to overfit to adversarial examples seen during training, therefore still being vulnerable to unknown adversaries. 

Given the current situation, Google Brain organized a NIPS 2017 competition that encouraged researchers to develop new methods to generate adversarial examples as well
as to develop new ways to defend against them~\cite{nipscompetition}. After the NIPS 2017 competition concluded and the solutions were all open-sourced, the CAAD 2018 competition was organized with a nearly identical ruleset, allowing competitors to push the state-of-the-art beyond the final NIPS 2017 results. In the CAAD 2018 competition, our solutions placed \textbf{1st} in both the \textit{non-targeted} and \textit{targeted} attack competitions, and \textbf{3rd} in the \textit{defense} competition. Detailing these solutions will be the core focus of this report. 

In section~\ref{sec:comp}, we detail the NIPS 2017 competition rules, overview top-placing solutions, and clarify the few differences between the NIPS 2017 and CAAD 2018 ruleset. In sections~\ref{sec:nontargeted},~\ref{sec:targeted}, and~\ref{sec:defense} we detail our final solutions and development results for the \textit{non-targeted} attack, \textit{targeted} attack, and \textit{defense} competitions, respectively. Finally, in section~\ref{sec:conc}, we conclude the paper with a brief discussion summarizing our findings. Our solutions have been open-sourced\footnote{\url{https://github.com/ysharma1126/caad\_18}}, and we hope the community find them useful.

\section{Competition Details} \label{sec:comp}
\subsection{NIPS 2017 Competition}
Most of the details provided can be found in the corresponding Springer book chapter from the ``NIPS 2017 Competition Book''. Please refer for further detail~\cite{nipscompetition}. 
\subsubsection{Dataset}
The competition organizers desired to use a dataset which 1) is large enough and nontrivial, 2) a well-known problem, so competitors can potentially reuse existing classifiers, and 3) has never been used before, so competitors are unlikely to overfit. 

These requirements were resolved by choosing the image classification problem and
creating a dataset with ImageNet-compatible images~\cite{deng2009imagenet}. To do this, the organizers collected a set of images which were never used in publicly available datasets, labelled them using pretrained ImageNet classifiers, and then manually verified that these labels were correct. This dataset was separated in two, the DEV dataset and the TEST dataset. 

The DEV dataset was released to participants at the beginning of the competition, so
they can use it for development. The TEST dataset was kept secret and was used to evaluate final submissions of the
participants. The DEV dataset and TEST dataset contained 1000 and 5000 images, respectively.

\subsubsection{Rules}

As discussed, the competition had three sub-tracks, non-targeted attacks, targeted attacks, and defenses. The submitted programs were given a set of images (and target classes in case of targeted attack) as an input, and had to produce either an adversarial image (for attack submission) or classification label (for defense submission) for each input image. Attacks were also constrained, both in perturbation size and computational resource usage, to ensure the distortions were visually imperceptible and the generation process was computationally practical. 
\vspace{-3mm}
\subparagraph{Perturbation Size}
An unnormalized integer $L_{\infty}$ difference between 4 and 16, inclusive. Value chosen at test-time by organizers and provided as an argument to the submitted programs. 
\vspace{-3mm}
\subparagraph{Computation} 
Submissions have to process a batch of 100 images in 500 seconds (including any time required for initialization and setup) on a machine with 24GB of RAM and a NVIDIA Tesla K80 GPU. 
\subsection{NIPS 2017 Solutions}
\subsubsection{Attacks}
\paragraph{Momentum Iterative Method (MIM)}
The Fast Gradient Sign Method (FGSM) is a one-step adversarial attack which calculates the gradient only once using the assumption of linearity of the decision boundary about the data point. This attack is relatively successful, at least in the non-targeted case, suggesting that DNN decision boundaries are locally linear~\cite{goodfellow2014explaining}. However, this linear assumption may not hold when distortions are large. This is not a desirable property from a white-box attacker's perspective, as the adversarial examples generated ``underfit'' the model, limiting attack strength. 

In contrast, the basic iterative method (BIM) greedily moves the adversarial example in the direction of the gradient each iteration~\cite{kurakin2016adversarial}. This is not a desirable property from a black-box attacker's perspective, as the adversarial examples generated can easily ``overfit'' the model, limiting transferability. 

In order to alleviate this trade-off between attack strength and transferability, the \textbf{1st} place solution in both the non-targeted and targeted attack NIPS 2017 competitions integrated momentum into BIM for the purpose of stabilizing update directions and escaping from poor local optima~\cite{dong2017momentum}. They applied this momentum iterative method (MIM) to attack an ensemble of models, as an intrinsic direction that always fools multiple models is more likely to transfer to other models.

A set of models were selected primarily based on the estimated likelihood that they will be in the defense's solution. Given the computational resource restriction, ensembling more models would require performing less iterations to meet the time constraint. The team observed \textit{no transferability in the targeted case} and thus decided to, in this case, perform more iterations against several commonly used models, to maximize the white-box success rate. 

The submitted solution was not relatively unique, most participants used iterative gradient-based optimization against an ensemble of models. However, the novel integration of momentum seemed to provide the boost to push this solution over-the-top. Still, it is worth noting that the \textbf{1st} place targeted attack was not able to achieve transferability under the computational constraint, relying upon maximizing white-box success on classifiers most defenders used for their solutions. This clear shortcoming was what we primarily aimed to resolve. 

\paragraph{Adversarial Transformation Networks (ATN)}
One solution which did not use iterative gradient-based optimization at test-time and was successful, placing 4th in the non-targeted attack competition, instead leveraged \textit{adversarial transformation networks}, training fully-convolutional networks (FCNs) that can convert clean examples to adversarial examples~\cite{baluja2017atn}. 

Instead of doing test-time optimization, the solution is to simply train a neural network to learn to generate an adversarial example given an input. Therefore, at test-time, the network can be run on any input $x$ without requiring further access to the classifiers used for training or additional gradient computations, making it even faster than single-step gradient-based approaches, such as FGSM. This is particularly useful when under test-time computational constraints, which is the situation the NIPS 2017 competition organizers imposed upon attackers.

Using an ensemble of 8 classifiers, and all 13 possible integer perturbation sizes (4,5,...,16), the non-targeted FCN was trained. As mentioned, the team ranked 4th, demonstrating the effectiveness of the approach. In addition, the generated attack images were observed to have interesting properties: detailed textures are canceled out, and Jigsaw-puzzle-like patterns are added. These properties resulted in deceiving all tested classifiers to classifying the generated adversarial examples as the ``Jigsaw puzzle'' class, regardless of the input. The advantageous properties along with the desirable avoidance of the computational constraints this solution provides caused us to pay particular attention to it when preparing our submission. 

\subsubsection{Defenses}

As discussed, the sole defense which has not been completely broken/bypassed, and thus accepted to provide some increase in robustness is adversarial training. However, adversarial training on ImageNet is computationally impractical, the resources required are immense. In addition, the released adversarially trained networks are available to the entire community, and are therefore available to the attackers as well. Thus, defenders had to find alternative solutions in order to robustify existing classifiers.  
\paragraph{Guided Denoiser}
A natural idea for defense is to denoise adversarial examples before sending them to the target model. This approach, however, has been shown to suffer from an ``error amplification'' effect, where small residual perturbations are amplified to a large magnitude as a result of the forward propagation procedure~\cite{liao2017guided}. 

To solve this problem, the \textbf{1st} place solution in the defense competition set the loss function as the difference between the outputs of the target model induced by the original and adversarial examples, instead of the pixel-level reconstruction loss used in standard denoisers. Training using adversarial examples generated using BIM against 14 different ensembles with perturbation level $\epsilon=16$, they observe that their solution can yield a robustness improvement using significantly less training data and time than adversarial training, implying that solely learning to denoise is easier than learning the coupled task of classification and defense~\cite{liao2017guided}. Though it has been found that the guided denoiser is not robust in the white-box case~\cite{athalye2018cvpr}, the NIPS 2017 competition performance demonstrated its black-box robustness to existing transfer attacks.  

\paragraph{Randomized Preprocessor} The guided denoiser solution was unique, in fact, most solutions attempted to use preprocessors to robsutify existing ImageNet classifiers. Many solutions used an ensemble of preprocesors and an ensemble of classifiers, but surprisingly, the best ``preprocessor + classifier'' solution, placing \textbf{2nd}, used a single preprocessor and a single classifier. The core trait of this solution was the utilization of randomization. 

As a preprocessor, a random resizing layer and a random padding layer were prepended to the strongest released ensemble adversarially trained classifier~\cite{xie2017randomization, tramer2017ensemble}. Importantly, the amount of resizing was constrained to  be within a reasonably small range, in order to not cause the network performance on clean images to significantly drop, demonstrating the trade-off between clean accuracy and adversarial robustness ``preprocessor-based defenses'' inevitably encounter. At inference time, prediction results are averaged over 30 randomization patterns for each image. Though, alike the guided denoiser, it's been shown that this defense is not robust in the white-box case~\cite{athalye2018obfuscated}, the NIPS 2017 competition performance evidenced its black-box robustness to existing transfer attacks.  

\subsection{CAAD 2018 Ruleset}

The CAAD 2018 ruleset is remarkably similar to the NIPS 2017 competition, allowing participants to iterate on the open-sourced NIPS 2017 solutions. The relevant differences were in regards to perturbation size and computational resource usage. 
\vspace{-3mm}
\subparagraph{Perturbation Size}
A fixed unnormalized integer $L_{\infty}$ difference of 32. Known to the participants in the development stage, and double the maximum posssible perturbation size allowed in the NIPS 2017 competition.
\vspace{-3mm}
\subparagraph{Computation} 
Submissions have to process a batch of 1000 images in 1000 seconds (including any time required for initialization and setup) on a machine with 24GB of RAM and a NVIDIA Tesla P100 GPU. 

In summary, the CAAD 2018 competition format was identical to that of the NIPS 2017 competition, except the perturbation constraint was relaxed while the computation constraint was strengthened. This certainly influenced the design of our solutions, which we will delve into in the subsequent sections.

\section{Non-Targeted Attack}
\label{sec:nontargeted}
\subsection{Algorithm}

Though offset by the relaxed perturbation size constraint and the higher-end GPU, the computational time constraint was significantly strengthened. Solutions in the CAAD 2018 competition need to process input images nearly \textbf{5x} faster than NIPS 2017 solutions. 

Using iterative optimization-based attacks, such as the aforementioned top-placing NIPS 2017 solution, MIM, the strengthened computational constraint forces a significant reduction in the ensemble size and/or the number of iterations. As in the non-targeted case, transferability was readily possible, maximizing ensemble size was prioritized over maximizing number of iterations. Therefore, keeping the ensemble fixed (8 models), the number of iterations had to be decreased from 10 to 7 to meet the time constraint. Submitting the top-placing NIPS 2017 solution with the iteration reduction (using the increased perturbation size) to the development round, our submission placed 7th, indicating that a significant improvement upon previous work was required. 

At this point, we realized the aforementioned advantages of adversarial transformation networks, and how they could be used to take advantage of the significant increase in perturbation size while avoiding the more stringent computational time constraint. 

With this insight, we developed a very simple solution. Using an adversarial transformation network trained with maximum $\epsilon=16$, we generate an adversarial example with perturbation size $\epsilon=16$. We then simply scale up the magnitude of the perturbation by 2x, to take advantage of the CAAD 2018 maximum $\epsilon=32$. Unlike the targeted case, where the adversarial perturbation needs to be carefully crafted in order to achieve the target classification, the larger the perturbation, the better in the non-targeted case. Confident in our strategy, we benchmarked the solution. 

\subsection{Results}

\begin{table}[!h]
\centering
\begin{tabular}{c|c|c|c|c|c|c|c}
 & IncV3 & AdvResV2 & NASNet & Guided & Randomized & 2 MSB & Dropout \\
 \hline
 Clean & 3.1\% & 3.5\% & 1.9\% & 1.7\% & 2.9\% & 10.7\% & 22.2\%\\
 FGSM (IncV3) & 72.8\% & 15.2\% & 33.9\% & 18.6\% & 14.6\% & 26.7\% & 46.2\% \\
  MIM (8) & 100\% & 98.5\% & 92.9\% & 69.9\% & 71.5\% & 60.4\% & 60.8\%\\
  ATN & \textbf{100\%} & \textbf{99.9\%} & \textbf{100\%} & \textbf{100\%} & \textbf{99.9\%} & \textbf{98.0\%} & \textbf{66.7\%} \\
\hline
\end{tabular}
\caption{Non-Targeted Attack Success Rate (ASR) using 1000 DEV examples}
\label{tab:nontarget}
\end{table}

For benchmarking, we use all 1000 images in the DEV dataset. For evaluation, we used existing open-sourced ImageNet classifiers, cleanly trained and adversarially trained, the top NIPS 2017 defense solutions, and two novel defenses we proposed in the context of the competition. For non-targeted attacks, we tested a weak attack, single-step FGSM against a single model, the top NIPS 2017 solution, MIM against an ensemble of 8 models (7 iterations), and our proposed adversarial transformation network (ATN) solution. Please see the results in Table~\ref{tab:nontarget}. 

The first 3 columns denote existing classifiers. ``IncV3'' is the cleanly trained Inception-V3 classifier~\cite{szegedy2015going}, a baseline model white-box attacked in the NIPS 2017 examples and used in all NIPS 2017 attack solutions (including the NIPS 1st place MIM attack and our ATN solution evaluated here). ``AdvResV2'' is the strongest released adversarially trained model, ensemble adversarially trained Inception ResNet-V2~\cite{szegedy2016resnet}, used in many defense solutions and also ensembled in the MIM and ATN attack solutions. ``NASNet'' is the recent state-of-the-art cleanly trained ImageNet classifier found via neural architecture search~\cite{zoph2017nas}, used as a proxy for non-targeted transfer to cleanly trained networks out of ensemble.  

The next 2 columns, ``Guided'' and ``Randomized'' are the 1st and 2nd place NIPS 2017 defense solutions, respectively. Note that the ``Guided'' submitted solution used 4 models, each with an associated denoiser, with 2 out of the 4, ensemble adversarially trained Inception-V3 and Inception ResNet-V2, included in the MIM and ATN ensembles. In addition, the ``Randomized'' submitted solution also uses ensemble adversarially trained Inception ResNet-V2. 

The final 2 columns are two novel defenses we developed for the defense track of this competition. First, ``2 MSB'' simply passes the 2 most significant bits (MSBs) of the input example to all 5 released adversarially trained models\footnote{Downloaded from \url{https://github.com/tensorflow/models/tree/master/research/adv_imagenet_models}} for classification. As the defender knows the maximum perturbation is $L_{\infty}=32$, preserving only the 2 MSBs will remove the perturbation. This immense preprocessing will inevitably hurt the clean accuracy, however, but as its been observed that adversarially trained networks perform feature selection to yield robustness, it was hypothesized that these networks would be more robust to the preprocessing~\cite{tsipras2018lunch}.

Finally, ``Dropout'' strengthens the ``Randomized'' defense by instead of solely doing random resizing and padding, thereby only affecting pixels on the border of the image, also performs random dropout ($p=0.1$), using smoothing to fill in the selected ``dropped out'' values. In addition, instead of only using the ensemble adversarially trained Inception ResNet-V2 model, which most attackers will be white-box attacking, ensemble it with ``NASNet''.

As can be seen, the two novel ``stronger'' defenses yield a significant reduction in clean accuracy, with ``Dropout'' having a clean accuracy of 77.8\%. This reduction is reflected in relatively poor performance against the weak FGSM adversary. However, against the stronger MIM winning solution, these defenses are significantly more accurate than the top NIPS defenses, demonstrating yet again the ``clean-adversarial accuracy tradeoff'' encountered by preprocessor-based defenses.

The MIM attack is relatively successful, achieving 90+\% ASR against existing ImageNet classifiers, even against `NASNet' which is out-of-ensemble, and performing relatively well against the top NIPS defenses and even the strengthened defenses. Nonetheless, ATN performs significantly better, achieving near 100\% ASR against all evaluation models (including the top NIPS defenses), except ``Dropout''. However, given the poor clean accuracy of ``Dropout'', yielding relatively poor performance against weak adversaries, such strong preprocessors are unlikely to be submitted as defense solutions. Regardless, ATN outperforms MIM against ``Dropout'' as well. 

This significant improvement can be explained by the fact that the ATN solution avoids being constrained by the competition rules. As MIM, and other iterative gradient-based optimizers, perform optimization at test-time on each given input, very few iterations could be performed. However, as ATN is trained offline, significantly more ``training'' can be done, yielding stronger perturbations. This advantageous property of ATN, performing the optimization offline and thus being faster than even single-step FGSM at test-time, proves to be quite useful in such situations. 

\section{Targeted Attack}
\label{sec:targeted}
\subsection{Algorithm}

For targeted attacks, we found that a successfully trained ATN was not available for use, so as we did not have the resources for training, we resorted to improving upon the NIPS 1st place MIM solution. Recall that all targeted attack NIPS 2017 solutions were not able to transfer, and thus resorted to maximizing white-box success rate against commonly used models. Even with the increased perturbation size, we found that this continued to be true, iterative gradient-based optimization on an ensemble of classifiers did not yield targeted transferability. 

For transferability, one needs to find perturbations which induce the desired effect against unknown models. In the targeted case, this is significantly harder, as the gradient directions which point towards the specified target class need to be robust to transfer. It is not enough to simply induce misclassification. 

A natural solution to increasing robustness is to optimize through randomization. Ideally one could simulate the transfer attack situation, randomly sampling models from a large functional distribution. Unfortunately, due to the limited set of high-performing models and the limited size of GPU memory, this approach is infeasible. On the other hand, optimizing through randomized preprocessing certainly is feasible. Inspired by~\cite{xie2018diversity}, we test optimizing through the random resizing and padding preprocessing performed in the ``Randomized'' defense. We also tested optimizing through ``Dropout'', but were unsuccessful, possibly due to its output range being larger, making it a harder optimization problem. 

Another natural solution is to regularize the gradients during optimization, thereby encouraging the resulting perturbations to be more generalizable. In the ideal case, this would result in gradients pointing in directions salient to human perception~\cite{tsipras2018lunch}. To do this simply, we performed spatial smoothing on the gradient using a gaussian filter, choosing a gaussian filter due to its advantageous properties and theoretical justifications. 

\subsection{Results}

\begin{table}[!h]
\centering
\begin{tabular}{c|c|c|c|c|c|c}
 & IncV3 & AdvResV2 & NASNet & Guided & Randomized & Dropout \\
 \hline
 MIM (1000) & 100\% & 96.4\% & 0.52\% & 0\% & 0.84\% & 0\%\\
 Rand (10) & 73.5\% & 17.9\% & 0.86\% & 0.05\% & 17.6\% & 0\%\\
 Rand (100) & 97.4\% & 47.8\% & 42.5\% & 16.2\% & 46.4\% & 4.13\% \\
  Rand (500) & 100\% & 82.2\% & 50.3\% & 43.8\% & 82.9\% & 11.5\%\\
  Rand (1000) & 100\% & 91.7\% & 71.5\% & 68.9\% & 90.5\% & 22.0\% \\
\hline
\end{tabular}
\caption{Targeted ASR using 1000 DEV examples. Numbers in parentheses indicate the number of iterations the optimization was run for.}
\label{tab:target_1}
\end{table}

In our first experiments, we aimed to achieve targeted transferability on ImageNet using the randomization strategy, disregarding the computational constraint.  ``MIM'' is the NIPS 2017 1st place targeted attack solution, which includes ``IncV3'' and ``AdvResV2'' in the ensemble. ``Rand'' is MIM except instead of simply optimizing on the input example, each input example is preprocessed with the ``Randomized'' preprocessor each iteration. Note that ``Rand'' is technically white-box attacking the ``Randomized'' defense. Please see the results in Table~\ref{tab:target_1}.

The NIPS 1st place solution used 20 iterations, due to the computational constraint. Even if the number of iterations is increased to 1000, targeted transferability is not yielded. However, using randomized preprocessing, significant targeted transferability can be yielded with at least 100 iterations. Yet, note the poor white-box ASR against the adversarially trained network. This significant reduction in white-box ASR can be improved, along with transferability, with more iterations, reaching 90+\% with 1000 iterations. 

However, under the NIPS 2017 computational constraint, ensemble MIM can only do a maximum of 20 iterations; 100, let alone 1000 iterations is in drastic violation of that. Under the strengthened CAAD 2018 computational constraint, ensemble MIM can only do a maximum of \textbf{10 iterations}. Seeing the poor results with ``Rand (10)'', the randomization strategy is clearly not a solution worthy of submission under the computational constraint. 

\begin{table}[!h]
\centering
\begin{tabular}{c|c|c|c|c|c|c|c}
 & IncV3 & AdvResV2 & NASNet & Guided & Randomized & 2 MSB & Dropout \\
 \hline
 MIM & 99.7\% & \textbf{87.4\%} & 0.12\% & 0\% & 0.05\% & 0\% & 0\%\\
 Rand & 73.5\% & 17.9\% & 0.86\% & 0.05\% & 17.6\% & 0\%& 0\%\\
 Grad & \textbf{99.8\%} & 82.4\% & 29.6\% & \textbf{62.9\%} & 45.3\% & 0.04\%& 0\% \\
 Grad + Rand & 98.1\% & 72.7\% & \textbf{45.9\%} & 62.0\% & \textbf{58.4\%} & \textbf{0.15\%}& \textbf{0.10\%}\\
\hline
\end{tabular}
\caption{Targeted ASR using 1000 DEV examples and 10 iterations.}
\label{tab:target_2}
\end{table}

Given the computational constraint, we tested using spatial gradient smoothing via a gaussian filter with only 10 iterations. The results can be seen in Table~\ref{tab:target_2}. Surprisingly, unlike randomized preprocessing, gradient smoothing (``Grad'') yields a significant improvement in transferability without significantly harming the white-box success rate. In addition, by combining randomized preprocessing with gradient smoothing, transferability can be improved even further without a significant reduction in white-box success rate. The results using ``Grad + Rand'' are drastically better than the results using just ``Rand''. The unreasonable effectiveness of gradient smoothing is not well-understood and, given our results, is certainly a worthy subject for further study. 

For our final submission, we used ``Grad + Rand'' with a few modifications. First, the NIPS 2017 1st place ensemble was changed. Instead of ensembling Inception-V3 with 4 adversarially trained networks, we ensembled all 5 released adversarially trained networks, hypothesizing that defenses are not likely to use Inception-V3. In addition, in order to boost the white-box success rate, instead of performing randomized preprocessing every iteration (prob $p=1.0$), the randomization is performed half the time ($p=0.5$).

\section{Defense}
\label{sec:defense}
As discussed, we developed the ``2 MSB'' defense, which passes the 2 most significant bits (MSBs) of the input example to all 5 released adversarially trained models, and the ``Dropout'' defense, which in addition to random resizing and padding, performs random dropout ($p=0.1$), using smoothing to fill in the selected ``dropped out'' values, then passes the processed image to an ensemble of ``AdvResV2'' and ``NASNet''. Both yield a significant reduction in clean accuracy, with ``Dropout'' being worse, which is reflected in relatively poor performance against weak attacks, but also yield a significant improvement against the MIM NIPS 2017 winning solution, representative of strong attacks. We decided to submit ``2 MSB'' instead of ``Dropout'' to the defense track given its higher clean accuracy/weak attack performance, ``Dropout'' was simply too aggressive. However, ``Dropout'' performed much better than ``2 MSB'' against the ATN attack, which we were submitting. Rectifying this issue became our main focus for subsequent experiments. 

First, in order to increase general robustness, drawing inspiration from ``Dropout'' and ``Randomized'', we decided to test adding mild bernoulli random noise to the input, hoping the random noise distorts the crafted adversarial noise (``Noise''). In addition, recall that the perturbations generated by the trained ATN force classifiers to predominantly choose the ``Jigsaw puzzle'' class, regardless of input. Therefore, we tested modifying the defense to choose the 2nd most probable class if the most probable class is the ``Jigsaw puzzle'' class (``Avoid''). Our experimental results can be seen in Table~\ref{tab:defense}. 

\begin{table}[!h]
\centering
\begin{tabular}{c|c|c|c|c}
 & Dropout & 2 MSB & Noise & Noise+Avoid \\
 \hline
 Clean & 77.8\% & \textbf{89.3\%} & 85.1\% & 85.1\%\\
 FGSM (IncV3) & 53.8\% & \textbf{73.3\%} & 63.1\% & 63.1\%  \\
  MIM (8) & 39.2\% & 39.6\% & \textbf{44.2\%} & \textbf{44.2\%} \\
  ATN & 33.3\% & 2.0\% & 23.8\% & \textbf{42.0\%} \\
\hline
\end{tabular}
\caption{Defense Accuracy using 1000 DEV examples. ``Noise'' and ``Noise + Avoid'' build on the ``2 MSB'' solution.}
\label{tab:defense}
\end{table}

As can be seen, though ``Noise'' hurts the clean accuracy/performance against weak attacks, it still is quite better than ``Dropout''. In addition, ``Noise'' significantly improves performance against the strong attacks, both MIM and ATN. Using ``Avoid'', performance against ATN in particular is significantly improved, making the ``2 MSB'' solution significantly more robust than ``Dropout'' against our submitted attack, meriting submission. Note that this robustness improvement was only possible because the defense had perfect knowledge of the submitted adversary.

\section{Conclusion}
\label{sec:conc}
The CAAD 2018 competition took place with nearly identical rules to the NIPS 2017 competition, enabling participants to directly build upon previous solutions. For our \textit{non-targeted} attack solution, we used a trained \textit{adversarial transformation network} (ATN), allowing the optimization to be performed offline, thereby avoiding being limited by the more stringent time constraint. For our \textit{targeted} attack solution, we observed that transferability can be achieved by optimizing through randomization, though not efficiently enough to meet the given time constraint. Surprisingly, spatial gradient smoothing is able to achieve transferability while being efficient enough to meet the time constraint. In addition, by using spatial gradient smoothing, optimizing through randomization can improve transferability even further within the time constraint; the combination ended up being our final solution. For our \textit{defense} solution, we pass the 2 MSB of the input example to an ensemble of adversarially trained classifiers. We strenghen the defense even further by adding mild bernoulli random noise to the inputs, and patch the defense to improve its performance against our submitted non-targeted attack. Our \textit{non-targeted} and \textit{targeted} attack solutions both won \textbf{1st} place in their respective subtracks, and our \textit{defense} solution won \textbf{3rd} place. We hope our results can inform researchers in both generating and defending against adversarial examples. 

\bibliography{references}	
\bibliographystyle{plain}


\end{document}